\documentclass[conference]{IEEEtran}
\IEEEoverridecommandlockouts

\usepackage{cite}
\usepackage{amsmath,amssymb,amsfonts}
\usepackage{algorithmic}
\usepackage{graphicx}
\usepackage{textcomp}
\usepackage{xcolor}
\usepackage{times}
\usepackage{soul}
\usepackage{url}
\usepackage[hidelinks]{hyperref}
\usepackage[utf8]{inputenc}
\usepackage[small]{caption}
\usepackage{booktabs}
\usepackage{latexsym}
\usepackage{amsmath,amssymb,amsfonts,amsthm}
\usepackage{booktabs}
\usepackage{algorithm}
\usepackage{float}
\usepackage{multirow}

\renewcommand{\vec}[1]{\boldsymbol{#1}}

\def\BibTeX{{\rm B\kern-.05em{\sc i\kern-.025em b}\kern-.08em
    T\kern-.1667em\lower.7ex\hbox{E}\kern-.125emX}}

\begin{document}

\title{Toward Tag-free Aspect Based Sentiment Analysis: A Multiple Attention Network Approach}

\author{\IEEEauthorblockN{Yao Qiang}
\IEEEauthorblockA{\textit{Computer Science Department} \\
\textit{Wayne State University}\\
Detroit, USA \\
yao@wayne.edu}
\and
\IEEEauthorblockN{Xin Li}
\IEEEauthorblockA{\textit{Computer Science Department} \\
\textit{Wayne State University}\\
Detroit, USA \\
xinlee@wayne.edu}
\and
\IEEEauthorblockN{Dongxiao Zhu}
\IEEEauthorblockA{\textit{Computer Science Department} \\
\textit{Wayne State University}\\
Detroit, USA \\
dzhu@wayne.edu}
}

\maketitle

\begin{abstract}
	
Existing aspect based sentiment analysis (ABSA) approaches leverage various neural network models to extract the aspect sentiments via learning aspect-specific feature representations. However, these approaches heavily rely on manual tagging of user reviews according to the predefined aspects as the input, a laborious and time-consuming process. Moreover, the underlying methods do not explain how and why the opposing aspect level polarities in a user review lead to the overall polarity. In this paper, we tackle these two problems by designing and implementing a new Multiple-Attention Network (MAN) approach for more powerful ABSA without the need for aspect tags using two new tag-free data sets crawled directly from TripAdvisor (\url{https://www.tripadvisor.com}). With the Self- and Position-Aware attention mechanism, MAN is capable of extracting both aspect level and overall sentiments from the text reviews using the aspect level and overall customer ratings, and it can also detect the vital aspect(s) leading to the overall sentiment polarity among different aspects via a new aspect ranking scheme. We carry out extensive experiments to demonstrate the strong performance of MAN compared to other state-of-the-art ABSA approaches and the explainability of our approach by visualizing and interpreting attention weights in case studies.

\end{abstract}


\section{Introduction}

Customer reviews of products and services are increasingly available from e-commercial websites, blogs, micro-blogs, and online social media platforms, containing rich information that are valuable resources for both businesses and consumers. These reviews can help customers make purchasing decisions and choose restaurants and/or hotels. They can also help businesses become customer-centric, listening to their customers, understanding their voice, analyzing their feedback and learning more about customer experiences as well as their expectations for products or services. 

\begin{figure*}[h]
    \centering 
	\includegraphics[width=0.95\linewidth]{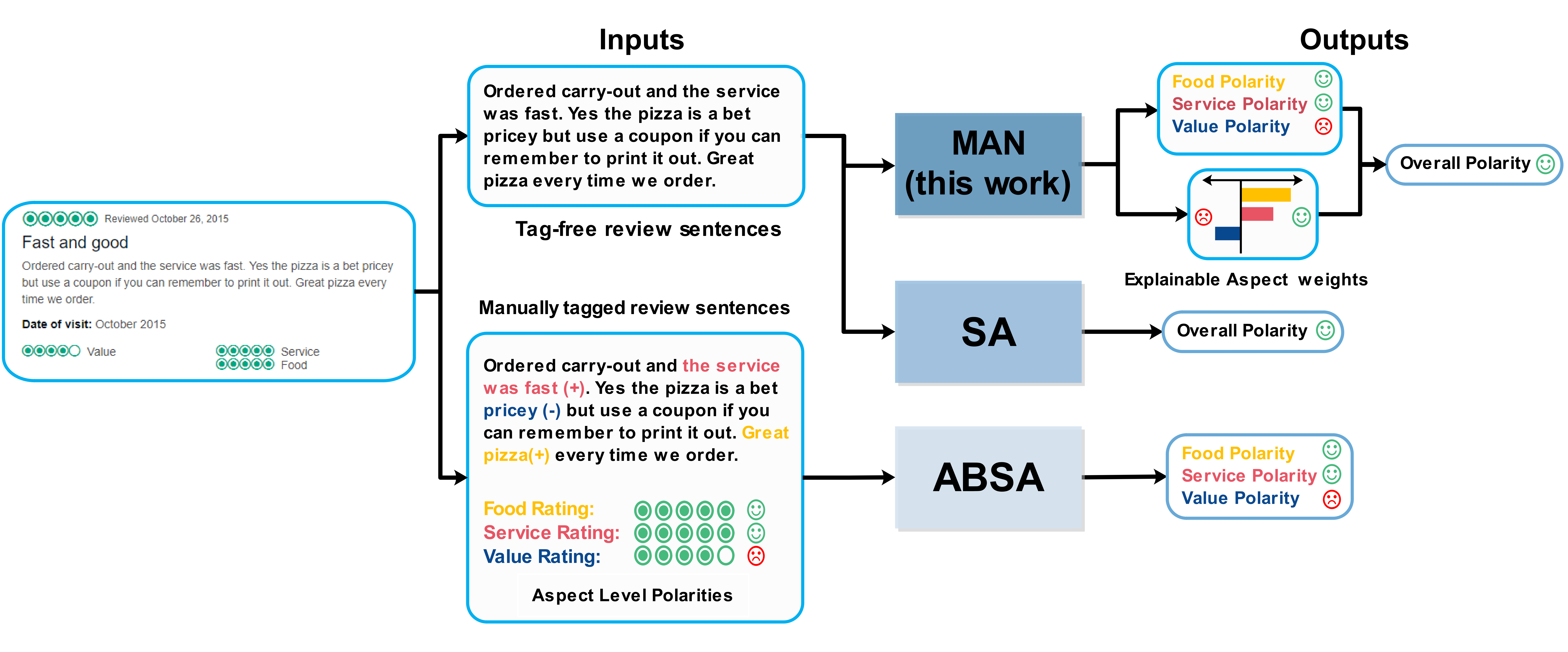}
	\caption{An overview of MAN-ABSA, SA and ABSA workflows.}   
	\label{fig:overview}
\end{figure*} 

Sentiment analysis (SA) has recently become a very active research area with the explosive increase of customer reviews due to its broad range of practical applications. Existing sentiment analyses are either performed at document level \cite{moraes2013document} or sentence level \cite{meena2007sentence}, both of which have been proven as highly efficient techniques in analyzing the myriad amount of texts. However, it is more common that customers may comment on a product or service from different aspects and express opposing polarities on these aspects as shown in the inputs of Figure \ref{fig:overview}. Aspect based sentiment analysis (ABSA) emerged as a more informative analysis to predict polarities based on the predefined aspect categories or aspect terms (tagged words/phrases) in the user reviews. This fine-grained trait of ABSA makes it a more effective application for businesses to monitor the ratio of positive to negative sentiment expressed towards specific aspects of a product or service, and extract valuable targeted insight. 


Nevertheless, most ABSA approaches are limited to a small number of academic data sets due to the lack of sufficiently tagged user reviews in terms of aspects. Although there are a huge number of customer reviews generated from e-commerce websites every day, manual tagging is a laborious and error-prone process, leading to the data sparsity challenge, which is often called cold-start problem \cite{song2019cold}, \cite{amplayo2018cold}. Moreover, many ABSA approaches are not fully explainable in terms of how the aspect polarities, often opposing, collectively lead to the overall polarity. To overcome these challenges, we propose a new Multiple-Attention Network (MAN) approach that directly exploits user-supplied tag-free review texts as well as aspect level and overall ratings crawled from review websites (e.g. TripAdvisor) to perform an automatic end-to-end ABSA. Using Self- and Position-Aware attention mechanism, MAN uncovers the weighted contribution of each word and phrase to aspect level and overall polarities. In addition, MAN can also detect the vital aspect(s) leading to the overall sentiment polarity among different aspects through a new aspect ranking scheme.


Figure \ref{fig:overview} illustrates the key differences between MAN, the existing ABSA, and SA approaches using a user review example from TripAdvisor. Our MAN approach takes the tag-free review sentences to infer the parts of sentence associated with aspect level and overall polarities, which can be considered as an automatic tagging approach supervised by overall and aspect level user ratings. The existing ABSA approaches typically analyze the manually tagged review sentences, e.g., ``the service was fast (+)", ``pricey (-)" and ``Great pizza (+)", corresponding to three pre-defined aspects $Service$, $Value$ and $Food$ whereas the existing SA approaches analyze the tag-free review sentences and simply infer an overall sentiment. 



We summarize our major contributions as follows: 

\begin{itemize}
	
	\item Combating the substantial data sparsity challenge in ABSA tasks, MAN automatically detects the vital words and phrases associated with aspect level and overall polarities without the need for manual tagging via a multiple-attention mechanism. 
	
	\item MAN is explainable in that the overall polarity can be attributed to aspect level polarities via a new aspect ranking scheme. To the best of our knowledge, this is among the first work to infer the relationship between aspect level and overall polarities to detect the most important aspect(s) of customer reviews. 
	
	\item We create two new data sets crawled directly from TripAdvisor with aspect level and overall ratings to demonstrate the potential for automatic deployment of MAN. These two data sets provide an objective and real-world evaluation of ABSA performance using tag-free review texts for this and future research. \footnote{\url{https://github.com/qiangyao1988/Toward-tag-free-ABSA}}
	
\end{itemize}

The rest of this paper is organized as follows. In Section 2, we review the related work on SA, ABSA, and attention mechanism. Our proposed model MAN is introduced in Section 3. In Section 4, we discuss the new tag-free data sets used in our experiments and report experimental results. We show the case study and visualization in Section 5. Finally, Section 6 concludes this paper and points out future research direction.

\section{Related Work}

\subsection{Sentiment Analysis}

The goal of sentiment analysis is to detect polarity in product or service reviews \cite{pang2008opinion}, \cite{liu2012sentiment}. Traditional classification algorithms, such as Naive Bayes \cite{liu2013scalable} and Support Vector Machine \cite{wang2012baselines}, are widely used for sentiment classification. These solutions mostly depend on feature engineering and manually defined rules, such as sentiment lexicon, n-grams, and dependency information. 


Neural network techniques without feature engineering are becoming increasingly popular for sentiment analysis. Classical neural network models are applied to solve this problem, such as Convolution Neural Network \cite{kim2014convolutional}, Recursive Neural Network \cite{dong2014adaptive} and Recurrent Neural Network \cite{mikolov2010recurrent}. More recently, Amplayo et al \cite{amplayo2018cold} designed a Hybrid Contextualized Sentiment Classifier model to capture the information from similar users/products and attempted to solve the cold-start problem in review sentiment classification. 


\subsection{Aspect Based Sentiment Analysis}

Aspect Based Sentiment Analysis can be divided into two subtasks: aspect extraction aims to identify aspects from the review sentences whereas aspect level sentiment classification is to detect aspect level polarities in review texts. 

In the aspect extraction subtask, traditional approaches extract frequent noun terms with dependency relations \cite{hu2004mining}, \cite{qiu2011opinion}. Other widely used approaches are based on supervised sequence labeling, such as Hidden Markov Model \cite{li2010structure} and Restricted Boltzmann Machine \cite{wang2012baselines}. To alleviate the need for large amounts of labeled data for training purposes, researchers developed an array of unsupervised approaches based on topic models, such as Latent Dirichlet Allocation \cite{blei2003latent} and its variants and extensions \cite{mukherjee2012aspect}, \cite{chen2013discovering} to extract and categorize aspects.

Similar to the sentiment classification problem, neural networks have been shown very powerful for the subtask of aspect level sentiment classification. Tang et al \cite{tang2015effective} used two single-direction LSTM's, called TD-LSTM, which combined the left context and right context together to detect aspects and classify sentiment polarities toward these aspects. Xue and Li \cite{xue2018aspect} is among one of the first works to employ a CNN-based model for aspect level sentiment analysis. More recently, Wang et al \cite{wang2019aspect} proposed a novel approach that models the specific segments for aspect level sentiment classification in a reinforcement learning framework. 


\subsection{Attention Mechanism}

Attention mechanism \cite{bahdanau2014neural}, a proven technique for semantic understanding of images and texts, allows the model to be attentive to aspect words with larger weights, or to deemphasize on words that are not relevant to any aspect. It was first introduced in a machine translation task, and since then has been extensively applied to solve image captioning \cite{xu2015show}, speech recognition \cite{chen2017recurrent} , recommender system \cite{ying2018sequential} and deep learning on biological data \cite{mahmud2018applications} problems. Recently, the attention mechanism has been used in a number of NLP tasks, such as language understanding \cite{shen2018disan} and question answering \cite{hermann2015teaching}. 

A variety of attention-based models have been introduced to tackle the problems in ABSA. Wang et al \cite{wang2016attention} applied attention and LSTM together in the model ATAE-LSTM by concatenating aspects with review word representations to compute the attention weights for aspect level sentiment classification. Ma et al \cite{ma2017interactive} designed an Interactive Attention Network (IAN) that used two attention networks to consider both attention on aspect and review context interactively. Song et al \cite{song2019attentional} proposed an Attentional Encoder Network (AEN), which avoids recurrence and employs attention based encoders for modeling context and aspect. Chen et al \cite{chen2019graph} treated the text as a graph and aspects as the specific areas of the graph to design a graph convolution neural networks and structural attention model for ABSA. Other models also emphasize that contextual word position information can also contribute to predicting the sentiment polarity of a specific aspect \cite{jiang2019position}.

Unfortunately, many of these ABSA models are designed for SemEval Task5 \cite{pontiki2016semeval} that rely on a few manually tagged user review data sets, which leads to the data sparsity challenge. More powerful ABSA approaches need to be developed in response to the real-world scenario that massive user-supplied review texts and ratings directly from e-commerce websites without laborious and error-prone manual tagging. Recent work in Song et al \cite{song2019cold} attempt to mitigate the cold-start problem by using the frequency-guided attention mechanism to accentuate on the most related targets, followed by composing the representations into a complementary vector for enhancing those of cold-start aspects. Nevertheless, the related targets used as complementary information are still from the small academic data sets in SemEval Task5. Therefore, new ABSA approaches accommodating the real-world tag-free data sets hold a strong promise to overcome the data sparsity challenge. Further, many existing ABSA approaches do not uncover the mechanism on how mixed aspect ratings lead to the overall rating. Attentive and hierarchical modeling architecture permitting dissection of the relationship between aspect level and overall polarities is highly desirable to identify the pivotal aspect(s) in the customer reviews.

\section{Model Description}

\subsection{Notations and Problem Formulation}

The online customer reviews of products or services can be represented as $\vec{R}$ = \{$\vec{r_1}$, $\vec{r_2}$, \ldots, $\vec{r_n}$\}. Each review consists of a number of words, i.e., $\vec{r}$ = (w$_1$, w$_2$, \ldots, w$_j$, \ldots, w$_m$), $n$ and $m$ denote the number of reviews and the number of words in each review, respectively. Besides the review sentences, there are often two types of ratings, i.e., the overall ratings and aspect ratings. Customers are not only allowed to give an overall rating but also encouraged to rate from different aspects, which are usually divided into several pre-defined broad categories in each domain. For example, in the restaurant domain, the four pre-defined aspect categories are \{$Food, Service, Value, Atmosphere$\} whereas in the hotel domain, they are \{$Room, Location, Value, Cleanliness$\}. In each review $\vec{r}$, the overall rating summarizes the customer's overall polarity $P^o$ towards the rated item, and the aspect ratings reflect the aspect level polarities $P^k$, where $k\in \left[ K \right] $ represents the index of the aspects, $\left[ K \right] := \{ 1,\ldots,K \}$, and $K$ denotes the number of the aspect categories. Our MAN-ABSA approach aims at predicting the overall polarity $P^o$ by analyzing the aspect level polarities $P^k$ in the review sentences.

\subsection{Model Architecture}

\begin{figure}[t]
	\includegraphics[width=0.5\textwidth]{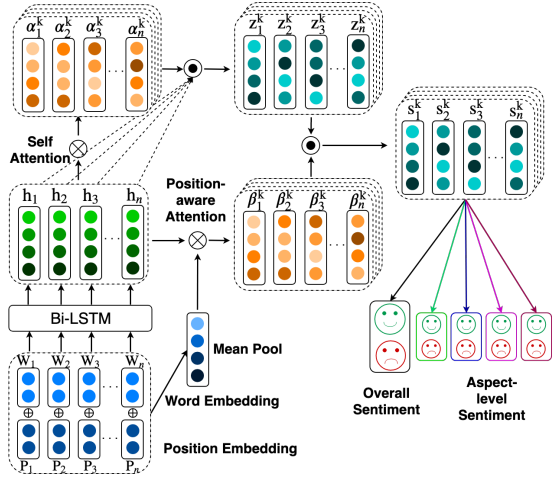}
	\caption{The MAN model architecture}
	\label{Figure2}
\end{figure}

As shown in Figure \ref{Figure2}, our MAN model consists of four modules, i.e., Embedding, Hidden States, Attention Encoder, and Sentiment Classifier. First, in the Embedding module, the review sentences are embedded into word vectors to capture low-level semantics, and the absolute positions of contextual words are embedded into the same embedding dimension of the word vectors to capture position information of the review texts\cite{shaw2018self}. Second, the Hidden States module takes the initially represented review texts via word and position embeddings as inputs to capture high-level semantics, which is the further representation of the review texts. Third, the Attention Encoder module takes the further representation and exploits multiple-attention mechanisms to capture the vital words leading to the aspect level polarities ($P^k$). Finally, in the Sentiment Classifier module, we concatenate the context vectors generated by different aspect attention encoders to get the final representations of the whole review sentences and use a fully connected layer to project the concatenated vector into the space of the target sentiment polarity classes ($P^o$). After the Sentiment Classifier module, we employ a model-agnostic aspect ranking scheme to identify the pivotal aspect(s) for the overall polarity in user reviews based on attention weights generated in the Attention Encoder module. More specifically, we describe the mathematics operations in each module below.


\subsection{Embedding Module}

After pre-processing the raw review texts, we employ the global word representation approach GloVe \cite{pennington2014glove} to obtain the fixed-length word embedding vector of each word. We map each word $w_j$ $\in$ $\mathbb{R}^{|V|}$ to its corresponding embedding vector $\vec{e_j}$ $\in$ $\mathbb{R}^{d\times 1}$, where $d$ is the word vector dimension (e.g. 300 in our experiments) and $V$ is the vocabulary size of the review data sets, respectively. Words beyond the GloVe vocabulary are randomly initialized with a uniform distribution U(-0.25, 0.25). Furthermore, in order to capture position information in the Attention Encoder module, we embed the absolute positions of context words into the same dimension $d$ through position embedding. The position embedding can model the different weights of words with different distances by converting the absolute position index to the embedding \cite{gu2018position}. The position embedding is randomly initialized and updated during the training process. In the end, we concatenate the word embedding with the position embedding in new embedding vectors $\vec{E}$ $\in$ $\mathbb{R}^{2d \times{V}}$. 


\subsection{Hidden States Module}

The Hidden States module aims to capture higher-level semantics using RNN networks. LSTM \cite{hochreiter1997long} and bidirectional variant (BiLSTM) are strong choices for further representations of the review texts because they have a built-in memory mechanism, which is designed in analogy to the psychological foundation of the human memory. Extracting more information from both past and future sequences of words, BiLSTM is more powerful than the standard LSTM. Given the embedding vector of each word $\vec{E}$ generated by the Embedding module into the Hidden States module, MAN generates the higher-level semantic representations of the review texts, i.e., $\vec{H}$ = $\text{LSTM}$ ($\vec{E}$) = ($\vec{h_1}$, $\vec{h_2}$, \ldots, $\vec{h_i}$, \ldots, $\vec{h_n}$), where $n$ denotes the size of the hidden states.  

\subsection{Attention Encoder Module}

The Attention Encoder module is applied to capture important information contributing to each aspect level polarity, which consists of two submodules: Self-Attention and Position-aware Attention. We can think of the Attention Encoder module as a two-step process. First, in the Self-Attention submodule, we filter the context words through a transformation matrix, which is able to detect the vital words contributing to the aspect level polarities. Then we capture the relevance of the filtered words through the Position-aware Attention submodule. 

\subsubsection{Self-Attention}

Given the high-level semantic representations $\vec{H}$, each output element generated by the Self-Attention submodule is computed as a weighted sum of a linearly transformed input elements $\boldsymbol{z_i} = \sum_{i=1}^n \alpha_i \boldsymbol{h_i}$.
The weight coefficient $\alpha_i$ is calculated using a softmax function: 
\begin{flalign} 
\alpha_i  = \frac {\exp \boldsymbol{f_i}}{\sum_{j=1}^n \exp \boldsymbol{f_j}} .  	
\end{flalign}
And $f_i$ is computed using a compatibility function of the hidden states: 
\begin{flalign} 
\boldsymbol{f_i} = \tanh (\boldsymbol{h_i} \boldsymbol{W}_\alpha \boldsymbol{h_i}^T + b_\alpha) ,   	
\end{flalign}
where $\boldsymbol{W}_\alpha$ $\in$ $\mathbb{R}$$^{n \times n}$ is a matrix mapping between the hidden states $\vec{h_i}$ and its transposition $\vec{h_i}^T$ and learned through the training process, $b_\alpha$ is a learnable bias. 

\subsubsection{Position-aware Attention}

Considering the contextual words that are closer to the aspect words may have a greater effect on the sentiment polarity \cite{jiang2019position}, we encode position information in the review texts using the Position-aware Attention submodule to capture the relevance of the contextual words. Given the output elements generated by the Self-attention submodule, the Position-aware Attention submodule computes a new weighted sum of linearly transformed elements $\boldsymbol{s_i} = \sum_{i=1}^n \beta_i \boldsymbol{z_i}$.
Similar to the self-attention, we can calculate the weight coefficient $\beta_i$ through a softmax function:
\begin{flalign} 
\beta_i = \frac {\exp \boldsymbol{g_i}}{\sum_{j=1}^n \exp \boldsymbol{g_j}}, 	
\end{flalign}
where $g_i $ is computed using a compatibility function of the hidden states transposition and the average of the word and position embedding: 
\begin{flalign} 
\boldsymbol{g_i} = \tanh (\boldsymbol{\bar{h}} \boldsymbol{W}_\beta \boldsymbol{h_i}^T + b_\beta) ,   
\end{flalign}
where $\vec{\bar{h}}$ is calculated by averaging the word and position embeddings, which captures both the global context of the review sentence \cite{he2017unsupervised} and the position information. $\boldsymbol{W}_\beta$ $\in$ $\mathbb{R}$$^{n \times n}$ is a training matrix mapping between the average embedding $\vec{\bar{h}}$ and the transposition $\vec{h_i}^T$, $b_\beta$ is a learnable bias. 

\subsection{Sentiment Classifier Module}

Following the Attention Encoder module, we proceed to the Sentiment Classifier module to get the aspect level and overall sentiment polarities. We concatenate results from the Attention Encoder module of different aspects $\boldsymbol{s^k}$ as the final overall representations $\boldsymbol{s^o} = \left[\boldsymbol{s}^1;\boldsymbol{s}^2;\ldots;\boldsymbol{s}^k\right]$. Then our model employs a fully connected layer to project the vectors $\boldsymbol{s^k}$ and $\boldsymbol{s^o}$ into the space of the target classes. In this paper, we set the number of classes $C=2$ corresponding to positive and negative. Formally: 
\begin{flalign} 
    y^k &= \mathrm{softmax} (\boldsymbol{W}_k\boldsymbol{s}^k + b_k) , \\
    y^o &= \mathrm{softmax} (\boldsymbol{W}_o\boldsymbol{s}^o + b_o) ,
\end{flalign}
where $y^k$ and $y^o$  $\in$ $\mathbb{R}^C$ is the predicted sentiment polarity probability distribution. $\boldsymbol{W}_k$, $\boldsymbol{W}_o$, $b_k$, and $b_o$ are learnable weight matrix and bias, respectively. 


\subsection{Regularization Term}

As different aspects have different attention encoders in our model, we add two regularization terms to make the sets of attention weights differentiate from each other to avoid the overlap of these attention encoders, in other words, the regularization terms diversify the attention weights from different aspects. We utilize orthogonal regularization similar to He et al \cite{he2017unsupervised}: 
\begin{flalign} 
\boldsymbol{M}_\alpha &= \left[\boldsymbol{\alpha}^1;\boldsymbol{\alpha}^2;\ldots;\boldsymbol{\alpha}^k\right] , \\
\boldsymbol{M}_\beta &= \left[\boldsymbol{\beta}^1;\boldsymbol{\beta}^2;\ldots;\boldsymbol{\beta}^k\right] , \\
\mathcal{R}_\alpha &= ||\boldsymbol{M}_\alpha^\mathrm{T}\boldsymbol{M}_\alpha - \boldsymbol{I}||_2 ,  \\	
\mathcal{R}_\beta &= ||\boldsymbol{M}_\beta^\mathrm{T}\boldsymbol{M}_\beta - \boldsymbol{I}||_2 ,  	
\end{flalign}
where $\boldsymbol{M_\alpha}$ and $\boldsymbol{M_\beta}$ contain the Self-Attention and Position-aware Attention weights, and $\boldsymbol{I}$ is the identity matrix. The two regularization terms $\mathcal{R}_\alpha$ and $\mathcal{R}_\beta$ reach the minimum value when the dot product between the different attention weights is zero. Then these regularization terms encourage orthogonality among the rows of the attention score matrix $\boldsymbol{M}_\alpha$ and $\boldsymbol{M}_\beta$ and penalize redundancy between different aspects vectors. 

\subsection{Objection Function} 

Our model is trained in an end-to-end manner by minimizing the cross-entropy loss between the target sentiment polarity and the predicted sentiment polarity for the review sentences as follows: 
\begin{flalign} 
    \mathcal{L}_k  &= -\hat{y}^k \log (y^k) - (1-\hat{y}^k)\log (1-y^k) ,\\
    \mathcal{L}_o  &= -\hat{y}^o \log (y^o) - (1-\hat{y}^o)\log (1-y^o) ,
\end{flalign}
where $\hat{y}^k$ and $\hat{y}^o$ are the aspect level and overall target sentiment polarities, respectively.

Furthermore, as our approach is multi-task learning \cite{li2018leveraging}, the losses from each task are combined to represent the training objectives of the entire model. In addition, in the real-world customer review data sets, some reviews may not mention all aspect categories in the review sentences. As shown in Figure \ref{fig:overview}, the customer only comments on and rates three out of four aspects. In this situation, the number of aspects changes adaptively in the training process. We set a hyperparameter $\delta$, which represents the number of aspects underlined in review texts. As our model is data-driven, we experiment with different $\delta$ ($\delta \in \left[ K \right] $) in two data sets to find the optimal value.

In this situation, the final objection function consists of the overall cross-entropy loss, the adaptive aspect level cross-entropy loss, the orthogonal regularization terms, and the L2 regularization term. Formally: 
\begin{flalign} 
    \mathcal{L}   = \mathcal{L}_o + \lambda _1 \sum_{k=1}^\delta \mathcal{L}_k + \lambda _2 \mathcal{R}_\alpha + \lambda _3 \mathcal{R}_\beta +  \lambda _4 ||\boldsymbol{\theta}|| ^2 ,
\end{flalign}
where $\lambda_1$, $\lambda_2$, $\lambda_3$, and $\lambda_4$ are tuning parameters to leverage the relative importance of different loss components, and $\boldsymbol{\theta}$ is the trainable parameter.

\subsection{Aspect Ranking Scheme}

In the Attention Encoder module, each word in the user review texts is associated with Self-Attention weight $\alpha_i$ and Position-aware Attention weight $\beta_i$ contributing to the sentiment polarities. In order to detect the vital aspect leading to the overall sentiment polarity among different aspects, we add up the two weights from each aspect and get a summarized importance score for each word corresponding to the specific aspect, i.e., 
\begin{flalign}  \label{{eq:summaryscore}} 
\text {score} ^k = \sum_{j=1}^m (\alpha ^k_j + \beta ^k_j) ,
\end{flalign}
where $\alpha ^k_j$ and $\beta ^k_j$ denote the attention scores of the $j_{th}$ word corresponding to $k_{th}$ aspect and the score$^k$ is the summarized importance score of the $k_{th}$ aspect in the review text. With this score, we employ an aspect ranking scheme to find the most important aspect(s) of the reviews, represented by higher summary scores.

\section{Experiments}

We perform extensive experiments to evaluate the performance of our MAN model assuming the overall polarity can be inferred from the different aspect level polarities, and the latter differentially influence the former as shown in Figures \ref{fig:overview} and \ref{Figure2}. In order to demonstrate the capability of our MAN approach in detecting vital words and explaining aspect level and overall polarities from rated customer reviews automatically with minimum external supervision, we create two new data sets from two different domains (restaurant and hotel) crawled directly from TripAdvisor including tag-free customer review texts and polarities represented by the overall rating and aspect ratings (1-5 stars). 

\subsection{Experimental Preparation}

\begin{figure} [h]
    \centering 
	\includegraphics[width=0.45\textwidth]{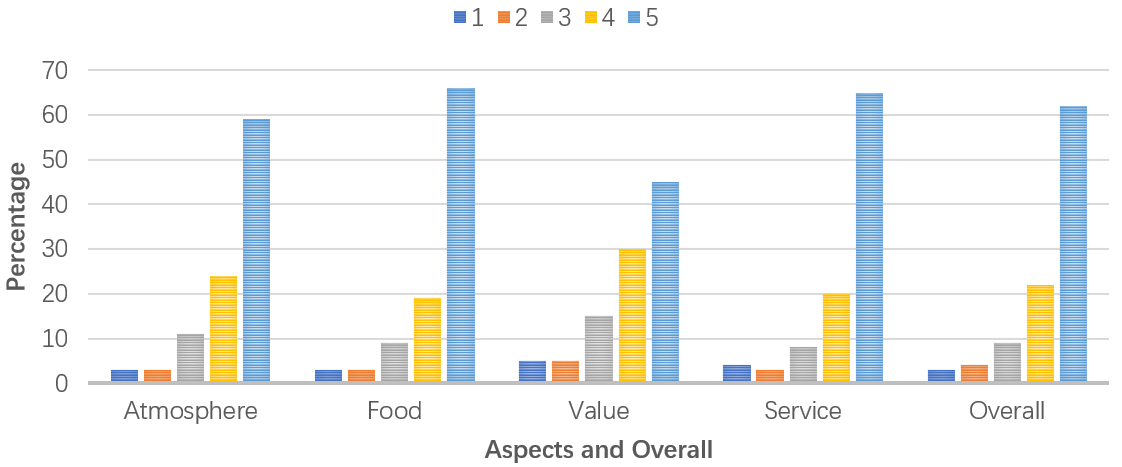}
	\caption{Distribution of ratings in the Restaurant data set}
	\label{Figure3}
\end{figure}

\begin{figure} [h]
    \centering 
	\includegraphics[width=0.45\textwidth]{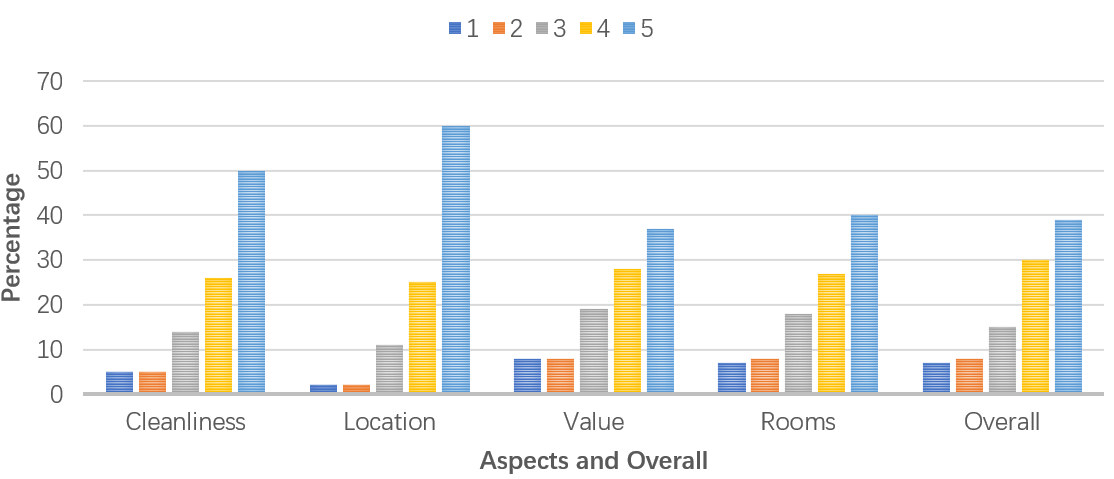}
	\caption{Distribution of ratings in the Hotel data set}
	\label{Figure4}
\end{figure}

\noindent{\bf Datasets}: In our experiments, using Scrapy (\url{https://scrapy.org}) framework, we crawl large numbers of restaurant and hotel reviews from TripAdvisor without manual processing and tagging. Table \ref{Table1} shows the summary statistics of the two data sets, including the number of reviews and the sentence lengths. Figures \ref{Figure3} and \ref{Figure4} illustrate the distributions of aspect level and overall ratings in each data set, respectively.
\begin{table}[h]
	\centering 
	\caption{Dataset summary statistics}  
	\label{Table1} 	
	\resizebox{0.45\textwidth}{!}
	{
	    \begin{tabular}{|c|c|c|c|c|c|}  
		    \hline  
		    \textbf{Domain}&\textbf{\#Reviews}&\textbf{Max tokens}&\textbf{Min tokens}&\textbf{Average tokens}\\
		    \hline
		    Restaurant&39,599&1,025&1&68\\
		    \hline
		    Hotel&31,999&240&2&80\\
		    \hline
	    \end{tabular}
	}
\end{table}

\noindent{\bf Pre-processing} involves a series of techniques aiming to optimize the performance of the experiments. In the beginning, we clean the raw data sets by deleting short reviews with less than three words. Then similar to other NLP analyses \cite {sun2014empirical}, we employ several standard pre-processing techniques, such as lemmatization, stemming, stop word removal and tokenization. As the purpose of our work is to classify and explain the overall polarity of reviews, we convert the original user ratings into positive and negative polarities. In addition, in order to avoid the open machine learning problem of class-imbalance, we treat the 3-star as negative polarity based on the observation that most reviews with 3-star rating clearly express a negative sentiment. So the five-star rating scale is binarized as positive if the rating is 4 or 5, otherwise negative \cite{guan2016weakly}. In the end, we generate an absolute position index for each review sentence based on the positions of the context words.

\noindent{\bf Experiment settings}: Our models and other baseline methods are implemented in PyTorch version 1.2.0 on a desktop workstation with a GTX 1080 GPU. We split each data set into three parts with 60/20/20 ratios, i.e., 60\% as the training set, 20\% as the validation set, and 20\% as the test set.

\subsection{Baseline Methods}

To evaluate the performance of our MAN model, we select a number of state-of-the-art models as the baselines for performance comparison. All the baseline methods perform in a multi-task learning manner corresponding to predict both the aspect level and overall polarities.

\noindent\textbf{CNN-static}\cite{kim2014convolutional} is the first model using convolution neuron network on text classification task with pre-trained vectors from $word2vec$ \cite{mikolov2013efficient} and provides a very strong baseline for sentiment classification. Similar to \cite{kim2014convolutional}, we set the widths of filters to 3, 4, 5 with 100 features each.\\
\textbf{GCAE}\cite{xue2018aspect} proposes a model based on CNN and gating mechanism. The novel Gated Tanh-ReLU Units can selectively output the sentiment features according to the given review texts.\\
\textbf{Standard-LSTM}\cite{wang2015predicting} uses LSTM networks to capture the high-level semantic representations of the review sentences and use the last hidden layer as the final sentence representation and feed to a softmax function to estimate the probability of each sentiment label. Furthermore, we concatenate the last hidden states of different aspects as the final overall sentiment representation. \\
\textbf{Standard-BiLSTM}\cite{wang2015predicting} is a variation of LSTM. The model concatenates the two representations from two different directions of the review sentences in order to capture more information than the Standard-LSTM for the sentiment polarities classification. \\
\textbf{AT-LSTM}\cite{wang2016attention} adds an attention layer following the LSTM layer aiming to capture the vital words in response to the whole review sentences for better sentiment polarities classification.\\
\textbf{AT-BiLSTM} \cite{wang2016attention} designs an attention mechanism together with the BiLSTM layer to capture more useful information for the sentiment polarities classification. \\
\textbf{BERT-base} \cite{sun2019fine} takes the final state $\vec h$ of the firs token $[CLS]$ as the representation of the whole sequence review sentence and adds a simple softmax classifier on the top of BERT to predict the probability of sentiment polarities. 

\subsection{Results and Analysis}

Due to the stochastic nature of our experiments, e.g., random initialization of model parameters and training data is processed in a random order, we run the baseline methods and our models on each data set five times, and report the averaged accuracy and macro-F1 score with variance of the overall sentiment polarity prediction as shown in Table \ref{Table2}. Our MAN models obtain substantial accuracy and macro-F1 improvements, which demonstrate the power of techniques applied in our model. Compared with models without any attention mechanism, MAN improves performance through the multiple-attention mechanisms. With this design, our model can learn the representations of vital words contributing to aspect-level and overall sentiment classifications. As the two data sets are highly imbalanced that there are much more positive reviews than the negative ones (Figures \ref{Figure3} and \ref{Figure4}), the macro-F1 score is more convincing than the accuracy. Therefore, the great improvement of the macro-F1 score demonstrates that our MAN models are very effective.  


\begin{table} [h]
	\centering 
	\caption{Comparison of different methods on Restaurant and Hotel data sets. Evaluation metrics are accuracy and macro-F1 score. Best scores are in bold.} 
	\label{Table2}
	\resizebox{0.48\textwidth}{!}{
	    \begin{tabular}{|c|c|c|c|c|}
	        \hline  
	        \multirow{2}*{\textbf{Model}} & \multicolumn{2}{|c|}{\textbf{Restaurant}} &\multicolumn{2}{|c|}{\textbf{Hotel}}\\
	        \cline{2-5} 
		    &\multicolumn{1}{|c|}{\textbf{ACC}} &\multicolumn{1}{|c|}{\textbf{Macro-F1}}&\multicolumn{1}{|c|}{\textbf{ACC}} &\multicolumn{1}{|c|}{\textbf{Macro-F1}}\\
		    \hline
		    CNN-static\cite{kim2014convolutional}&82.84$\pm{0.70}$&49.64$\pm{0.99}$&61.99$\pm{0.50}$&50.93$\pm{0.40}$\\
		    GCAE \cite{xue2018aspect}&82.99$\pm{0.31}$&51.13$\pm{0.57}$&62.90$\pm{0.18}$&50.41$\pm{0.34}$\\
		    BERT-base \cite{sun2019fine}&83.98$\pm{0.33}$&47.01$\pm{0.87}$&68.48$\pm{0.26}$&41.65$\pm{0.84}$\\
		    \hline
		    Standard-LSTM \cite{wang2015predicting}&82.59$\pm{1.06}$&54.03$\pm{0.48}$&63.81$\pm{0.08}$&53.40$\pm{0.17}$\\
		    Standard-BiLSTM \cite{wang2015predicting}&82.96$\pm{0.70}$&50.41$\pm{0.47}$&61.12$\pm{0.42}$&56.25$\pm{0.19}$\\
		    \hline
		    AT-LSTM \cite{wang2016attention}&81.05$\pm{0.46}$&54.06$\pm{0.55}$&62.28$\pm{0.31}$&55.56$\pm{0.42}$\\
		    AT-BiLSTM \cite{wang2016attention}&81.94$\pm{0.24}$&57.87$\pm{0.54}$&62.98$\pm{0.23}$&57.04$\pm{0.23}$\\
		    \hline
		    MAN-LSTM&89.42$\pm{0.17}$&77.29$\pm{0.24}$&82.97$\pm{0.75}$&79.64$\pm{0.36}$\\ 
		    \textbf{MAN-BiLSTM}&\textbf{89.64}$\pm{\textbf{0.18}}$&\textbf{77.60}$\pm{\textbf{0.26}}$&\textbf{83.06}$\pm{\textbf{0.14}}$&\textbf{79.81}$\pm{\textbf{0.22}}$\\ 
		    \hline
	    \end{tabular}
	}
\end{table}


We also show the hyper-parameter tunning and the optimal selection in our experiments in Table \ref{Table3} and the results with different values of hyper-parameters $\delta$ in Table \ref{Table4}.

\begin{table} [h]
	\centering 
	\caption{Hyper-parameters tuning}  
	\label{Table3} 	
	\resizebox{0.45\textwidth}{!}{
	\begin{tabular}{|c|c|c|}  
		\hline  
		\textbf{Hyper-parameters}&\textbf{Values}&\textbf{Optimal Selection}\\
		\hline	
		Optimizer&SGD;ADAM&ADAM\\
		\hline
		Learning Rate&0.01;0.005;0.001&0.005\\
		\hline
		$\delta$&1;2;3;4&4\\
		\hline
		$\lambda_1$&0;0.2;0.5;1;1.5&0.5\\
		\hline
		$\lambda_2; \lambda_3$&0;0.1;0.2;0.5&0.5\\
		\hline
		$\lambda_4$&0;0.01;0.05&0.01\\
		\hline
	\end{tabular}
	}
\end{table}

\begin{table} [h]
	\centering 
	\caption{Comparison of the effect of $\delta$ on MAN.}  
	\label{Table4} 	
	\resizebox{0.45\textwidth}{!}{
	\begin{tabular}{|c|c|c|c|c|}
		\hline  
		\multirow{2}*{\textbf{$\delta$}} & \multicolumn{2}{|c|}{\textbf{Restaurant}} &\multicolumn{2}{|c|}{\textbf{Hotel}}\\
		\cline{2-5}
		&\textbf{ACC}&\textbf{Macro-F1}&\textbf{ACC}&\textbf{Macro-F1}\\
		\hline
		0&85.22$\pm{0.55}$&72.69$\pm{0.20}$&79.56$\pm{0.20}$&70.95$\pm{0.31}$\\
		\hline
		1&86.72$\pm{1.05}$&75.41$\pm{0.65}$&80.42$\pm{0.28}$&76.14$\pm{0.15}$\\
		\hline
		2&87.73$\pm{0.32}$&75.56$\pm{0.22}$&81.22$\pm{0.19}$&75.16$\pm{0.32}$\\
		\hline
		3&87.96$\pm{0.44}$&76.80$\pm{0.26}$&80.96$\pm{0.53}$&74.87$\pm{0.28}$\\
		\hline
		4&\textbf{89.64}$\pm{\textbf{0.18}}$&\textbf{77.60}$\pm{\textbf{0.26}}$&\textbf{83.06}$\pm{\textbf{0.14}}$&\textbf{79.81}$\pm{\textbf{0.22}}$\\
		\hline
	\end{tabular}
	}
\end{table}

\subsection{Ablation Study}

In this section, we show the experiment results of different variants of our model and examine the effectiveness of the key components (e.g. Position-aware attention mechanism and regularization terms) on the model performance via an extensive ablation study. As shown in Table \ref{Table5}, the performances of all the model ablations are inferior to our MAN model with regards to both accuracy and macro-F1 measures, which indicates that all of these discarded components are crucial for performance improvement. Comparing the results of MAN variants, the accuracy and macro-F1 measures of the MAN w/o Pos-Attention drop substantially on both data sets. It demonstrates the importance of contextual words position information, encoding the relevance between contextual words in review texts, on predicting the review sentiment polarities. In addition, the overall performance of MAN is also superior to all the variant models without the orthogonal regularization terms, which demonstrates the effectiveness to diversify the two different kinds of attention weights from different aspect attention encoders.  

\begin{table} [h]
	\centering 
	\caption{Ablation study of the MAN model} 
	\label{Table5}
	\resizebox{0.48\textwidth}{!}{
	\begin{tabular}{|c|c|c|c|c|}
		\hline  
		\multirow{3}*{\textbf{Model}} & \multicolumn{2}{|c|}{\textbf{Restaurant}} &\multicolumn{2}{|c|}{\textbf{Hotel}}\\
		\cline{2-5} 
		&\multicolumn{1}{|c|}{\textbf{ACC}} &\multicolumn{1}{|c|}{\textbf{Macro-F1}}&\multicolumn{1}{|c|}{\textbf{ACC}} &\multicolumn{1}{|c|}{\textbf{Macro-F1}}\\
		\hline
		MAN w/o Pos-Attention &83.96$\pm{0.53}$&70.28$\pm{0.32}$&81.21$\pm{0.47}$&75.57$\pm{0.27}$\\ 
		\hline
	    MAN w/o PA-Orth-Norm &89.04$\pm{0.18}$&75.46$\pm{0.14}$&81.85$\pm{0.77}$&77.03$\pm{0.44}$\\ 
		MAN w/o SA-Orth-Norm&88.24$\pm{0.34}$&74.87$\pm{0.32}$&81.58$\pm{0.28}$&76.59$\pm{0.21}$\\ 
		MAN w/o Orth-Norm&88.21$\pm{0.68}$&74.16$\pm{0.67}$&77.95$\pm{0.41}$&75.83$\pm{0.53}$\\ 
		MAN w/o L2-Norm&88.09$\pm{0.21}$&76.58$\pm{0.17}$&81.92$\pm{0.42}$&75.09$\pm{0.22}$\\ 
		\hline
		\textbf{MAN}&\textbf{89.64}$\pm{\textbf{0.18}}$&\textbf{77.60}$\pm{\textbf{0.26}}$&\textbf{83.06}$\pm{\textbf{0.14}}$&\textbf{79.81}$\pm{\textbf{0.22}}$\\ 
		\hline
	\end{tabular}
	}
\end{table}

\section{Case Study and Visualization}

\begin{figure}
	\includegraphics[width=0.50\textwidth]{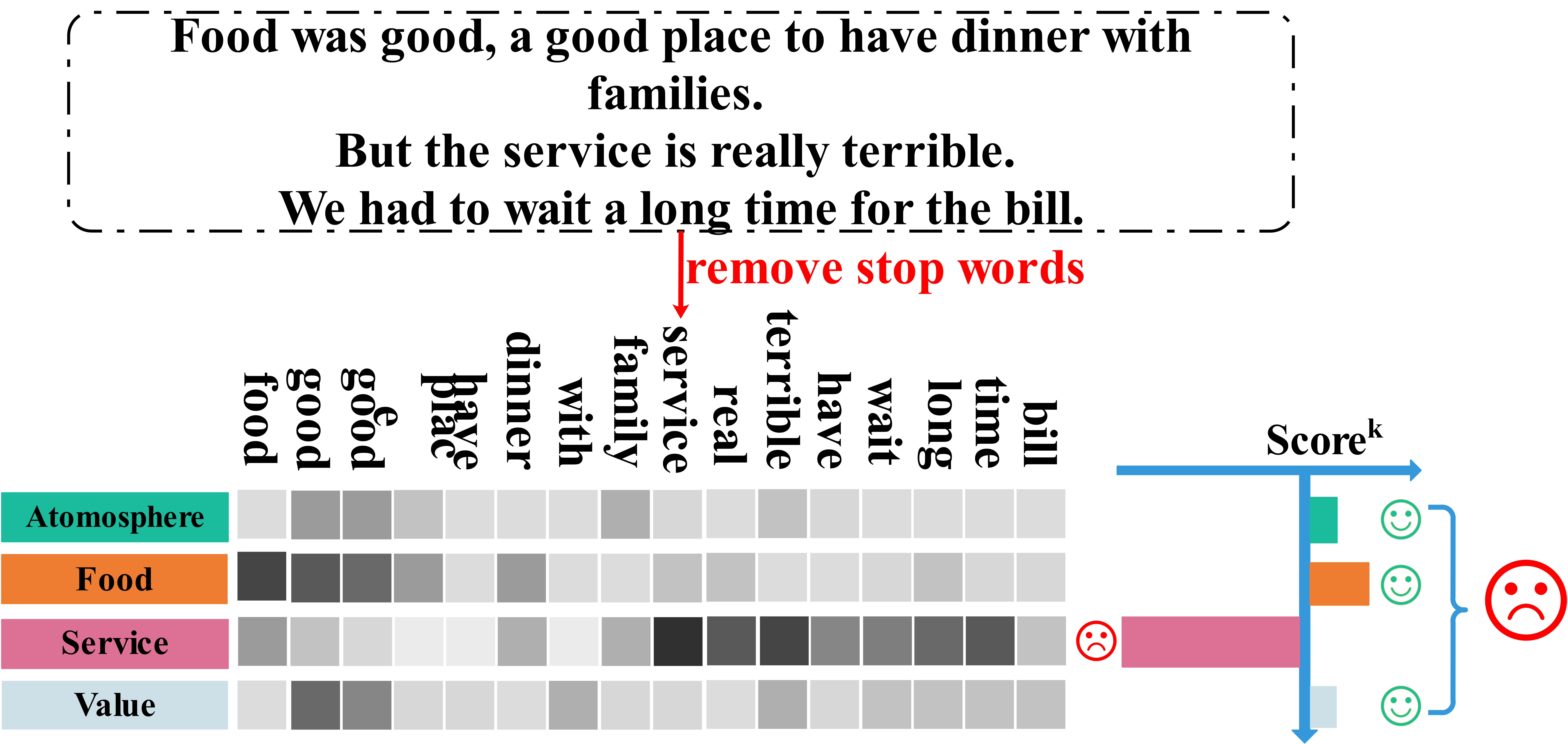}
	\caption{An example of opposing polarities in Restaurant review domain. (Best view in color.)}
	\label{Figure5}
\end{figure}

The performance evaluation of our MAN model demonstrates a more accurate overall polarity prediction on the test sets of the tag-free user reviews across both domains. To better understand its unique advantages over the existing approaches, particularly in explaining opposing polarities, we present a showcase example and interpret the result generated by our model (Figure \ref{Figure5}). 

In this example, we visualize the summation of the two different kinds of attention weights generated from the multiple-attention mechanism in our model for each word in the review text using a two-dimension grayscale heatmap. The shade of gray in each cell corresponds to the contribution of a particular word $w_j$ to the $k_{th}$ aspect polarity (attention weights), and the darker shade indicates more contribution. The length of horizontal bars (Score$^k$) corresponds to the magnitude of contributions of the entire sentence from each aspect to the overall polarity (Equation \ref{{eq:summaryscore}}). In this Restaurant review, the words ``food" and ``good" are the most important ones for positive polarity from $Food$ aspect whereas ``service", ``terrible", ``wait", ``long" and ``time" are the most important words for negative polarity from $Service$ aspect. The negative polarity from $Service$ aspect with a larger summary score dominates over the positive polarity from $Food$ aspect, leading to the overall negative polarity. Furthermore, we check the predicted sentiment polarity of this example among all the models we compared with, our MAN model is the only one that is capable of inferring the overall negative polarity correctly.

\section{Conclusions and Future Work}

In this paper, we develop a new MAN model to overcome the cold-start challenge in ABSA via utilization of the increasingly available aspect ratings. The Self-Attention is designed to detect the vital words contributing to the aspect level polarity, and the Position-Aware Attention is capable of capturing the relevance of the vital words by paying more attention to the closer words. Moreover, the two orthogonal regularization terms encourage the diversity of attention weights from different aspect attention encoders. The new model-agnostic aspect ranking scheme finds the most important aspect(s), which contributes most to the overall polarity, making the overall polarity more explainable, particularly with opposing aspect-level polarities. Using experimental studies, we demonstrate the superior performance of MAN compared with other state-of-art models in both polarity detection and explanation. Finally, we expect the MAN model trained on the tag-free user review data sets can be deployed to enable a fully automatic ABSA in the near future, opening a new line of ABSA research to combat the data sparsity challenge. 

For future work, as user-supplied text reviews are continuously available, it is desirable to expand the MAN model with continual learning techniques to automatically update sentiment polarities without laborious tagging of user reviews and re-training of the entire model. The idea of applying active learning to deal with the scarce labeled data \cite{nezhad2019deep} is another direction to deal with the cold-start problem in ABSA.

\clearpage
\bibliographystyle{IEEEtran}
\bibliography{IEEEabrv,cite}

\begin{thebibliography}{10}
\providecommand{\url}[1]{#1}
\csname url@samestyle\endcsname
\providecommand{\newblock}{\relax}
\providecommand{\bibinfo}[2]{#2}
\providecommand{\BIBentrySTDinterwordspacing}{\spaceskip=0pt\relax}
\providecommand{\BIBentryALTinterwordstretchfactor}{4}
\providecommand{\BIBentryALTinterwordspacing}{\spaceskip=\fontdimen2\font plus
\BIBentryALTinterwordstretchfactor\fontdimen3\font minus
  \fontdimen4\font\relax}
\providecommand{\BIBforeignlanguage}[2]{{%
\expandafter\ifx\csname l@#1\endcsname\relax
\typeout{** WARNING: IEEEtran.bst: No hyphenation pattern has been}%
\typeout{** loaded for the language `#1'. Using the pattern for}%
\typeout{** the default language instead.}%
\else
\language=\csname l@#1\endcsname
\fi
#2}}
\providecommand{\BIBdecl}{\relax}
\BIBdecl

\bibitem{moraes2013document}
R.~Moraes, J.~F. Valiati, and W.~P.~G. Neto, ``Document-level sentiment
  classification: An empirical comparison between svm and ann,'' \emph{Expert
  Systems with Applications}, vol.~40, no.~2, pp. 621--633, 2013.

\bibitem{meena2007sentence}
A.~Meena and T.~Prabhakar, ``Sentence level sentiment analysis in the presence
  of conjuncts using linguistic analysis,'' in \emph{ECIR}.\hskip 1em plus
  0.5em minus 0.4em\relax Springer, 2007, pp. 573--580.

\bibitem{song2019cold}
K.~Song \emph{et~al.}, ``Cold-start aware deep memory networks for multi-entity
  aspect-based sentiment analysis,'' 2019.

\bibitem{amplayo2018cold}
R.~K. Amplayo, J.~Kim, S.~Sung, and S.-w. Hwang, ``Cold-start aware user and
  product attention for sentiment classification,'' \emph{arXiv preprint
  arXiv:1806.05507}, 2018.

\bibitem{pang2008opinion}
B.~Pang, L.~Lee \emph{et~al.}, ``Opinion mining and sentiment analysis,''
  \emph{Foundations and Trends{\textregistered} in Information Retrieval},
  vol.~2, no. 1--2, pp. 1--135, 2008.

\bibitem{liu2012sentiment}
B.~Liu, ``Sentiment analysis and opinion mining,'' \emph{Synthesis lectures on
  human language technologies}, vol.~5, no.~1, pp. 1--167, 2012.

\bibitem{liu2013scalable}
B.~Liu, E.~Blasch, Y.~Chen, D.~Shen, and G.~Chen, ``Scalable sentiment
  classification for big data analysis using naive bayes classifier,'' in
  \emph{2013 IEEE international conference on big data}.\hskip 1em plus 0.5em
  minus 0.4em\relax IEEE, 2013, pp. 99--104.

\bibitem{wang2012baselines}
S.~Wang and C.~D. Manning, ``Baselines and bigrams: Simple, good sentiment and
  topic classification,'' in \emph{Proceedings of the 50th annual meeting of
  the ACL: Short papers-volume 2}.\hskip 1em plus 0.5em minus 0.4em\relax
  Association for Computational Linguistics, 2012, pp. 90--94.

\bibitem{kim2014convolutional}
Y.~Kim, ``Convolutional neural networks for sentence classification,''
  \emph{arXiv preprint arXiv:1408.5882}, 2014.

\bibitem{dong2014adaptive}
L.~Dong, F.~Wei, C.~Tan, D.~Tang, M.~Zhou, and K.~Xu, ``Adaptive recursive
  neural network for target-dependent twitter sentiment classification,'' in
  \emph{Proceedings of the 52nd annual meeting of the association for
  computational linguistics (volume 2: Short papers)}, 2014, pp. 49--54.

\bibitem{mikolov2010recurrent}
T.~Mikolov, M.~Karafi{\'a}t, L.~Burget, J.~{\v{C}}ernock{\`y}, and
  S.~Khudanpur, ``Recurrent neural network based language model,'' in
  \emph{Eleventh annual conference of the international speech communication
  association}, 2010.

\bibitem{hu2004mining}
M.~Hu and B.~Liu, ``Mining and summarizing customer reviews,'' in
  \emph{Proceedings of the tenth ACM SIGKDD international conference on
  Knowledge discovery and data mining}.\hskip 1em plus 0.5em minus 0.4em\relax
  ACM, 2004, pp. 168--177.

\bibitem{qiu2011opinion}
G.~Qiu, B.~Liu, J.~Bu, and C.~Chen, ``Opinion word expansion and target
  extraction through double propagation,'' \emph{Computational linguistics},
  vol.~37, no.~1, pp. 9--27, 2011.

\bibitem{li2010structure}
F.~Li \emph{et~al.}, ``Structure-aware review mining and summarization,'' in
  \emph{Proceedings of the 23rd international conference on computational
  linguistics}.\hskip 1em plus 0.5em minus 0.4em\relax Association for
  Computational Linguistics, 2010, pp. 653--661.

\bibitem{blei2003latent}
D.~M. Blei, A.~Y. Ng, and M.~I. Jordan, ``Latent dirichlet allocation,''
  \emph{Journal of machine Learning research}, vol.~3, no. Jan, pp. 993--1022,
  2003.

\bibitem{mukherjee2012aspect}
A.~Mukherjee and B.~Liu, ``Aspect extraction through semi-supervised
  modeling,'' in \emph{Proceedings of the 50th annual meeting of the ACL: Long
  papers-volume 1}.\hskip 1em plus 0.5em minus 0.4em\relax Association for
  Computational Linguistics, 2012, pp. 339--348.

\bibitem{chen2013discovering}
Z.~Chen, A.~Mukherjee, B.~Liu, M.~Hsu, M.~Castellanos, and R.~Ghosh,
  ``Discovering coherent topics using general knowledge,'' in \emph{Proceedings
  of the 22nd ACM international conference on Information \& Knowledge
  Management}.\hskip 1em plus 0.5em minus 0.4em\relax ACM, 2013, pp. 209--218.

\bibitem{tang2015effective}
D.~Tang \emph{et~al.}, ``Effective lstms for target-dependent sentiment
  classification,'' \emph{arXiv preprint arXiv:1512.01100}, 2015.

\bibitem{xue2018aspect}
W.~Xue and T.~Li, ``Aspect based sentiment analysis with gated convolutional
  networks,'' \emph{arXiv preprint arXiv:1805.07043}, 2018.

\bibitem{wang2019aspect}
T.~Wang, J.~Zhou, Q.~V. Hu, and L.~He, ``Aspect-level sentiment classification
  with reinforcement learning,'' in \emph{2019 International Joint Conference
  on Neural Networks (IJCNN)}.\hskip 1em plus 0.5em minus 0.4em\relax IEEE,
  2019, pp. 1--8.

\bibitem{bahdanau2014neural}
D.~Bahdanau, K.~Cho, and Y.~Bengio, ``Neural machine translation by jointly
  learning to align and translate,'' \emph{arXiv preprint arXiv:1409.0473},
  2014.

\bibitem{xu2015show}
K.~Xu \emph{et~al.}, ``Show, attend and tell: Neural image caption generation
  with visual attention,'' in \emph{International conference on machine
  learning}, 2015, pp. 2048--2057.

\bibitem{chen2017recurrent}
P.~Chen \emph{et~al.}, ``Recurrent attention network on memory for aspect
  sentiment analysis,'' in \emph{Proceedings of the 2017 conference on
  empirical methods in natural language processing}, 2017, pp. 452--461.

\bibitem{ying2018sequential}
H.~Ying, F.~Zhuang, F.~Zhang, Y.~Liu, G.~Xu, X.~Xie, H.~Xiong, and J.~Wu,
  ``Sequential recommender system based on hierarchical attention networks,''
  in \emph{the 27th International Joint Conference on Artificial Intelligence},
  2018.

\bibitem{mahmud2018applications}
M.~Mahmud \emph{et~al.}, ``Applications of deep learning and reinforcement
  learning to biological data,'' \emph{IEEE transactions on neural networks and
  learning systems}, vol.~29, no.~6, pp. 2063--2079, 2018.

\bibitem{shen2018disan}
T.~Shen, T.~Zhou, G.~Long, J.~Jiang, S.~Pan, and C.~Zhang, ``Disan: Directional
  self-attention network for rnn/cnn-free language understanding,'' in
  \emph{Thirty-Second AAAI Conference on Artificial Intelligence}, 2018.

\bibitem{hermann2015teaching}
K.~M. Hermann, T.~Kocisky, E.~Grefenstette, L.~Espeholt, W.~Kay, M.~Suleyman,
  and P.~Blunsom, ``Teaching machines to read and comprehend,'' in
  \emph{Advances in neural information processing systems}, 2015, pp.
  1693--1701.

\bibitem{wang2016attention}
Y.~Wang, M.~Huang, L.~Zhao \emph{et~al.}, ``Attention-based lstm for
  aspect-level sentiment classification,'' in \emph{Proceedings of the 2016
  conference on empirical methods in natural language processing}, 2016, pp.
  606--615.

\bibitem{ma2017interactive}
D.~Ma, S.~Li, X.~Zhang, and H.~Wang, ``Interactive attention networks for
  aspect-level sentiment classification,'' \emph{arXiv preprint
  arXiv:1709.00893}, 2017.

\bibitem{song2019attentional}
Y.~Song, J.~Wang, T.~Jiang, Z.~Liu, and Y.~Rao, ``Attentional encoder network
  for targeted sentiment classification,'' \emph{arXiv preprint
  arXiv:1902.09314}, 2019.

\bibitem{chen2019graph}
J.~Chen, H.~Hou, Y.~Ji, and J.~Gao, ``Graph convolutional networks with
  structural attention model for aspect based sentiment analysis,'' in
  \emph{2019 International Joint Conference on Neural Networks (IJCNN)}.\hskip
  1em plus 0.5em minus 0.4em\relax IEEE, 2019, pp. 1--7.

\bibitem{jiang2019position}
T.~Jiang \emph{et~al.}, ``A position-aware transformation network for
  aspect-level sentiment classification,'' in \emph{2019 International Joint
  Conference on Neural Networks (IJCNN)}.\hskip 1em plus 0.5em minus
  0.4em\relax IEEE, 2019, pp. 1--8.

\bibitem{pontiki2016semeval}
M.~Pontiki, D.~Galanis, H.~Papageorgiou, I.~Androutsopoulos, S.~Manandhar,
  A.-S. Mohammad, M.~Al-Ayyoub, Y.~Zhao, B.~Qin, O.~De~Clercq \emph{et~al.},
  ``Semeval-2016 task 5: Aspect based sentiment analysis,'' in
  \emph{Proceedings of the 10th international workshop on semantic evaluation
  (SemEval-2016)}, 2016, pp. 19--30.

\bibitem{shaw2018self}
P.~Shaw, J.~Uszkoreit, and A.~Vaswani, ``Self-attention with relative position
  representations,'' \emph{arXiv preprint arXiv:1803.02155}, 2018.

\bibitem{pennington2014glove}
J.~Pennington, R.~Socher, and C.~Manning, ``Glove: Global vectors for word
  representation,'' in \emph{Proceedings of the 2014 conference on empirical
  methods in natural language processing (EMNLP)}, 2014, pp. 1532--1543.

\bibitem{gu2018position}
S.~Gu, L.~Zhang, Y.~Hou, and Y.~Song, ``A position-aware bidirectional
  attention network for aspect-level sentiment analysis,'' in \emph{Proceedings
  of the 27th International Conference on Computational Linguistics}, 2018, pp.
  774--784.

\bibitem{hochreiter1997long}
S.~Hochreiter and J.~Schmidhuber, ``Long short-term memory,'' \emph{Neural
  computation}, vol.~9, no.~8, pp. 1735--1780, 1997.

\bibitem{he2017unsupervised}
R.~He, W.~S. Lee, H.~T. Ng, and D.~Dahlmeier, ``An unsupervised neural
  attention model for aspect extraction,'' in \emph{Proceedings of the 55th
  Annual Meeting of the Association for Computational Linguistics (Volume 1:
  Long Papers)}, 2017, pp. 388--397.

\bibitem{li2018leveraging}
X.~Li, D.~Zhu, and P.~Levy, ``Leveraging auxiliary measures: a deep multi-task
  neural network for predictive modeling in clinical research,'' \emph{BMC
  medical informatics and decision making}, vol.~18, no.~4, pp. 45--53, 2018.

\bibitem{sun2014empirical}
X.~Sun, X.~Liu, J.~Hu, and J.~Zhu, ``Empirical studies on the nlp techniques
  for source code data preprocessing,'' in \emph{Proceedings of the 2014 3rd
  International Workshop on Evidential Assessment of Software
  Technologies}.\hskip 1em plus 0.5em minus 0.4em\relax ACM, 2014, pp. 32--39.

\bibitem{guan2016weakly}
Z.~Guan \emph{et~al.}, ``Weakly-supervised deep learning for customer review
  sentiment classification.'' in \emph{IJCAI}, 2016, pp. 3719--3725.

\bibitem{mikolov2013efficient}
T.~Mikolov \emph{et~al.}, ``Efficient estimation of word representations in
  vector space,'' \emph{arXiv preprint arXiv:1301.3781}, 2013.

\bibitem{wang2015predicting}
X.~Wang, Y.~Liu, S.~Chengjie, B.~Wang, and X.~Wang, ``Predicting polarities of
  tweets by composing word embeddings with long short-term memory,'' in
  \emph{Proceedings of the 53rd Annual Meeting of the Association for
  Computational Linguistics and the 7th International Joint Conference on
  Natural Language Processing (Volume 1: Long Papers)}, vol.~1, 2015, pp.
  1343--1353.

\bibitem{sun2019fine}
C.~Sun, X.~Qiu, Y.~Xu, and X.~Huang, ``How to fine-tune bert for text
  classification?'' \emph{arXiv preprint arXiv:1905.05583}, 2019.

\bibitem{nezhad2019deep}
M.~Z. Nezhad, N.~Sadati, K.~Yang, and D.~Zhu, ``A deep active survival analysis
  approach for precision treatment recommendations: Application of prostate
  cancer,'' \emph{Expert Systems with Applications}, vol. 115, pp. 16--26,
  2019.

\end{thebibliography}
\end{document}